\pdfoutput=1

\documentclass[11pt]{article}

\usepackage[final]{acl}

\usepackage{times}
\usepackage{latexsym}

\usepackage[T1]{fontenc}

\usepackage[utf8]{inputenc}

\usepackage{microtype}

\usepackage{inconsolata}

\usepackage{graphicx}

\usepackage{lipsum}
\usepackage{tabularx}
\usepackage{booktabs}
\usepackage{makecell}
\usepackage{multirow}
\usepackage{multicol}
\usepackage{bbding}
\usepackage{colortbl}
\usepackage{xcolor}
\usepackage{underscore}
\usepackage[normalem]{ulem}
\usepackage{latexsym}

\title{EduFlow: Advancing MLLMs' Problem-Solving Proficiency through Multi-Stage, Multi-Perspective Critique}

\author{
  \textbf{Chenglin Zhu}\textsuperscript{1,2}\thanks{\ Equal contribution.} \quad
  \textbf{Tao Zhang}\textsuperscript{1}\footnotemark[1] \quad
  \textbf{Chong Li}\textsuperscript{1,2}\footnotemark[1] \quad
  \textbf{Mingan Lin}\textsuperscript{1}$^{\dagger}$ \quad
  \textbf{Zenan Zhou}\textsuperscript{1}$^{\dagger}$ \quad
  \textbf{Jian Xie}\textsuperscript{1}$^{\dagger}$  \\
  \textsuperscript{1}Baichuan Inc. \quad
  \textsuperscript{2}Peking University \\
  \texttt{\{zhuchenglin, lichong\}@stu.pku.edu.cn} \\
  \texttt{\{zhangtao.tanh, zenanchow\}@gmail.com} \\
  \texttt{\{linmingan, xiejian\}@baichuan-inc.com}
}

\begin{document}
\maketitle
\begin{abstract}

Multimodal large language models (MLLMs) still perform poorly on scientific tasks, particularly those requiring multi-step and interpretable reasoning. Their limitations include insufficient scientific reasoning patterns, lack of global coherence in multi-step inference, and the absence of reflective self-correction, making them unreliable in structured scientific contexts. We introduce \textbf{EduFlow}, the first end-to-end framework that covers the full pipeline of educational scientific reasoning, including data selection, MCTS-based trajectory construction, model training, and output optimization. At its core is \textbf{EduPRM}, a process-aware reward model that critiques reasoning steps with tags, and justifications. EduPRM is trained via curriculum learning on three complementary supervision sources: MCTS-guided trajectories, error-injected critiques, and teacher–student dialogues, enabling dynamic adaptation to multi-stage problem solving and iterative refinement during inference. We further propose \textbf{EduMCTS}, a domain-adapted search framework that introduces bootstrapping actions specifically designed for educational reasoning, such as a Self-Reflection mechanism that promotes reflective error correction. It further leverages EduPRM’s fine-grained feedback to guide the search toward higher-quality reasoning trajectories. By applying self-consistency and rejection sampling, we constructed \textbf{EduMCTS-160K}, a large-scale dataset of educational reasoning trajectories. Extensive experiments demonstrate that EduFlow enhances reasoning consistency and coherence. Code, data, and models will be released.
\end{abstract}

\section{Introduction}

\begin{figure}
    \centering
    \includegraphics[width=1\linewidth]{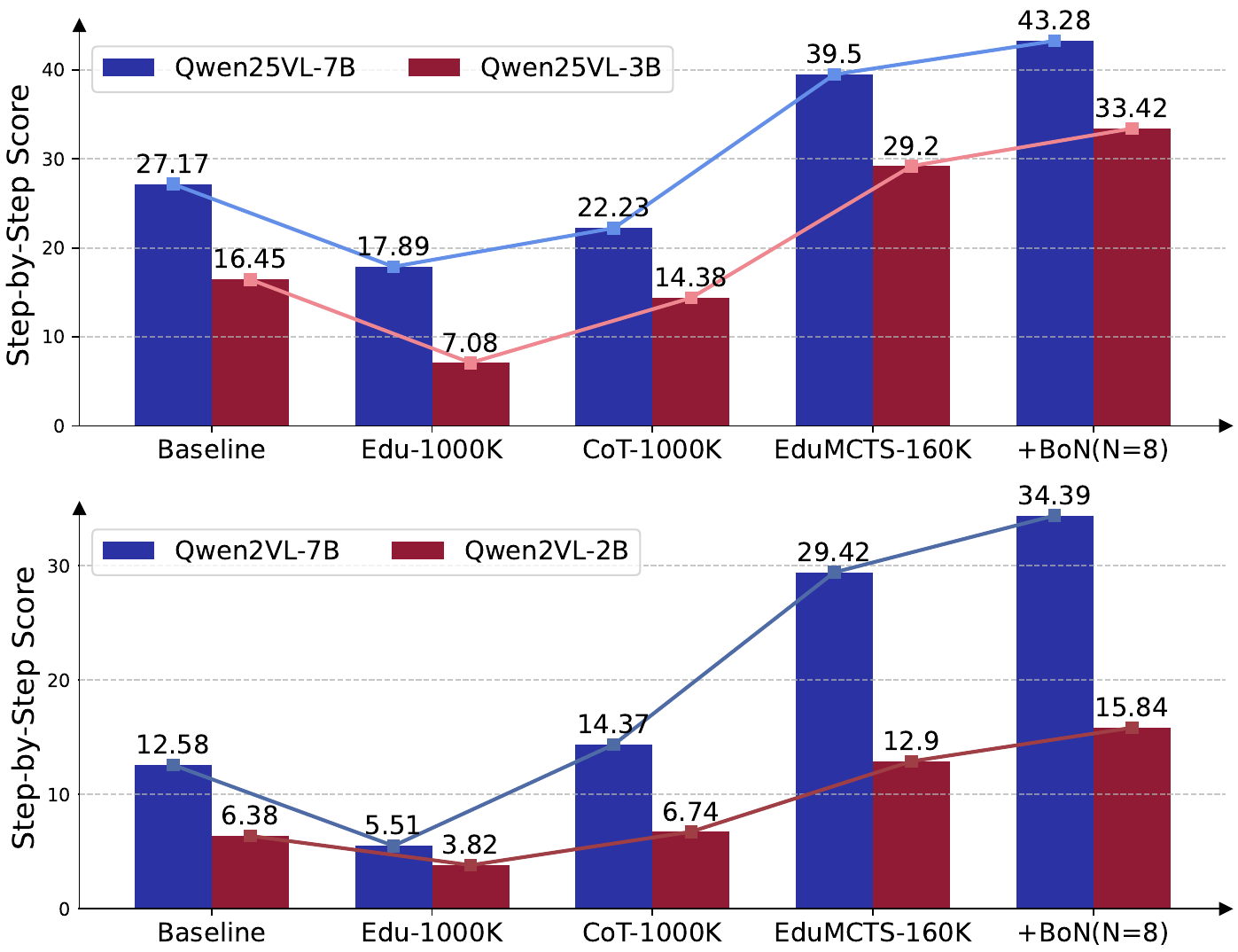}
    \caption{Step-by-step reasoning performance across different model sizes of the Qwen model family, training and inference strategies. We observe that each stage of EduFlow contributes to reasoning performance. BoN,N=8.}
    \label{fig:ablationstages}
\end{figure}

A striking revelation emerged from China’s 2024 National College Entrance Examination (Gaokao): while large language models (LLMs) achieved competitive scores in language-based subjects like Chinese and English, they catastrophically failed in mathematics, with none attaining even a passing grade~\citep{2023opencompass}.
This "subject imbalance" persists across modern benchmarks~\citep{K12Vista,zhou2025mdk12bench,zhang2024cmmmu, wang2024m4u, he2024cmmu, shen2024measuring}, where LLMs exhibit severe deficiencies in STEM disciplines.

The underperformance of LLMs in STEM domains stems from three key challenges. First, STEM-specific corpora—such as equations, symbolic logic, and structured problem-solving steps—are vastly underrepresented in pretraining stage compared to the abundance of open-domain texts from sources like literature, news, or social media. Second, reasoning errors accumulate because the model generates answers one token at a time without global planning or error checking, making multi-step reasoning especially fragile. Third, current models lack effective mechanisms for self-correction and reflection, making it difficult for them to move beyond memorized knowledge, or integrate novel insights in a coherent way. These limitations collectively hinder the ability of LLMs to perform rigorous, structured scientific inference.


To address these challenges, we introduce EduFlow. EduFlow provides a comprehensive solution that spans data selection, trajectory construction, model training, and output optimization—establishing a full-stack pipeline for enhancing reasoning quality in educational tasks.
It consists of three key stages, each tightly guided by EduPRM, which provides process-aware reward signals:
(1) PRM-guided data filtering, where EduPRM identifies flawed or unreliable reasoning trajectories from existing datasets by assigning low step-level rewards; these samples are then targeted for reconstruction via EduMCTS;
(2) PRM-guided data construction with EduMCTS, where EduPRM dynamically supervises the search process during Monte Carlo Tree Search to more efficiently generate diverse and high-reward trajectories;
(3) PRM-based best-of-N selection, where EduPRM scores and ranks multiple candidate outputs to select the most reliable final answer.
By unifying all stages under, EduFlow provides a generalizable and scalable blueprint for structured reasoning supervision in educational domains.


Process Reward Models (PRMs) have demonstrated superior performance over Outcome Reward Models (ORMs) in step-sensitive reasoning tasks~\citep{lightman2023lets,uesato2022solving},  yet existing approaches~\citep{liu2025mmc, guan2025rstarmath, lightman2023lets} face significant limitations: rigid data construction (e.g.,single-path annotations), scalar supervision signals that ignore LLMs’ generative capabilities, and static evaluation mechanisms incompatible with test-time scaling. To address these, we propose EduPRM, a process-aware reward model that critiques reasoning steps with scores, tags, and justifications. Unlike traditional PRMs, EduPRM integrates three complementary data strategies—MCTS-guided reasoning trajectories (automatically exploring diverse solution paths), error-injected critiques (simulating common student misconceptions), and teacher-student dialogues (refining ambiguous reasoning steps)—to capture multi-grained educational reasoning patterns. 

Although prior work has explored extending Process Reward Models PRMs to broader applications~\citep{wang2025visualprm, guan2025rstarmath, zhao2025genprm}, there remains no unified PRM framework that supports the entire optimization pipeline—from data construction to output selection. Trained via curriculum learning from atomic step annotations to complex reasoning decompositions, our proposed EduPRM dynamically adapts to multi-stage problem solving and enables iterative refinement during inference.

While chain-of-thought methods like LLaVA-CoT improve stepwise reasoning in general domains~\citep{xu2024llavacot}, they often plateau or degrade on science tasks due to limited trajectory diversity and lack of error correction. In contrast, MCTS offers a structured search paradigm for complex reasoning, but applying it to MLLMs faces two key challenges: (1) Ineffective Search: Without step-level guidance, MLLMs tend to explore homogeneous, low-quality paths, reducing search success~\citep{ren2025deepseekproverv2,zhang2024restmcts,guan2025rstarmath}; (2) High Cost: Multi-agent or tool-augmented variants are prohibitively expensive for large MLLMs~\citep{br2024vermcts,dong2024progressive,yao2024mulberry}. To address this, we propose EduMCTS, a domain-adapted framework that combines actor model generation with EduPRM-guided critique to iteratively construct high-quality reasoning paths. By integrating step-wise bootstrapping actions, step-level reward feedback, and self-reflection, EduMCTS produces pedagogically sound reasoning traces. The resulting dataset, EduMCTS-160K achives an 18\% higher success rate than LLaVA-CoT in science domains.

By integrating EduMCTS and EduPRM, our EduFlow framework achieves state-of-the-art performance across multiple educational benchmarks, with consistent improvements in both process-oriented (+8.2\%) and result-oriented (+8.2\%) metrics, while maintaining computational efficiency. Our key contributions are:
\begin{itemize}
\item \textbf{EduFlow:} the first end-to-end pipeline that unifies data selection, MCTS trajectory construction, model training and output optimization for educational scientific reasoning;
\item \textbf{EduPRM:} a process-aware reward model that critiques reasoning steps with scores, tags, and justifications, trained with three complementary supervision signals via curriculum learning to enable diverse scenarios;
\item \textbf{EduMCTS:} a domain-adapted search framework that integrates actor model generation, PRM-based stepwise guidance, bootstrapping actions, and self-reflective mechanisms to effectively navigate high-quality reasoning paths. This process yields \textbf{EduMCTS-160K}.
\item \textbf{Empirical validation:} extensive experiments show that EduFlow significantly narrows the gap between LLMs and human-level reasoning in complex educational tasks.
\end{itemize}

\section{Related Work}

\begin{figure*}[t]
\centering
\includegraphics[width=1.0\textwidth]{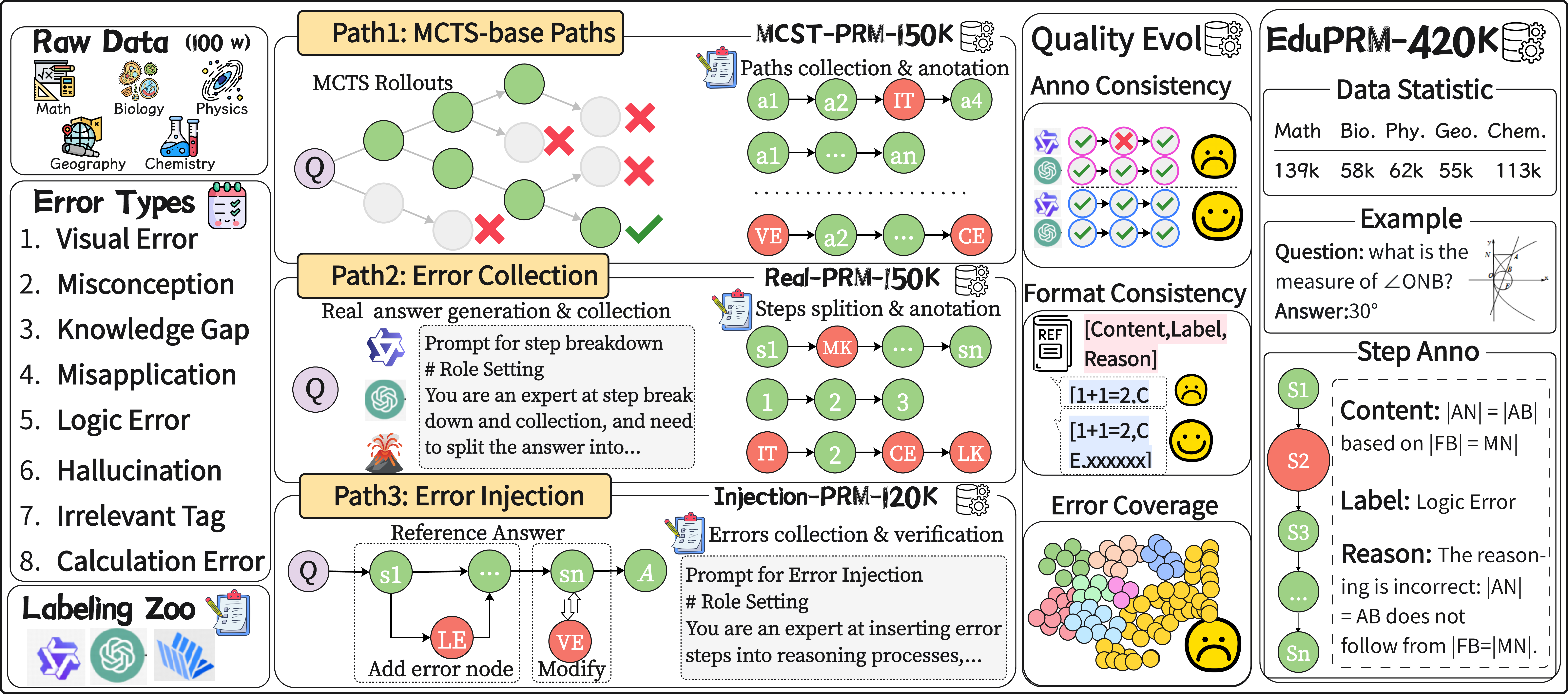} 
\caption{Overview of the construction and annotation pipeline for \textbf{EduPRM-420K}, a large-scale PRMs' dataset for educational multi-step reasoning. 
Raw data are collected from five STEM subjects (Math, Biology, Physics, Geography, Chemistry). 
Three complementary construction paths are designed: 
(1) \textbf{MCTS-base Paths};(2) \textbf{Error Collection}; (3) \textbf{Error Injection}. 
Each step is annotated with \texttt{Content}, \texttt{Label}, and \texttt{Reason}, covering 8 error types. 
We further ensure annotation quality through consistency checks, format validation, and error-type diversity metrics.
}
  \label{fig:prmpipeline}
\end{figure*}

\subsection{Edu Data}


Despite progress in multimodal LLMs, their performance on science remains limited~\citep{zhou2025mdk12bench,zhang2024cmmmu}. Models struggle with symbolic sparsity, fragile multi-step inference, and a lack of process-level feedback, especially in STEM subjects like physics and chemistry. Although recent benchmarks offer detailed evaluation protocols~\citep{he2024cmmu, shen2024measuring},, high-quality, stepwise training data is scarce due to challenges in collecting structured supervision, ensuring annotation consistency, and scaling long-form CoT generation.To address this, we construct Edu-1000K, a large-scale corpus of model-generated trajectories in educational domains. We then apply a multi-stage pipeline involving PRM-based selection, MCTS-guided generation, and refinement through self-consistency and rejection sampling, yielding EduMCTS-160K—a high-quality dataset of stepwise, pedagogically aligned reasoning traces.


\subsection{Process Evaluation}
Process reward models (PRMs) have shown strong performance in mathematical reasoning by providing step-level supervision~\citep{lightman2023lets, uesato2022solving}, but scaling them remains challenging due to costly annotation and limited data construction methods. While recent works introduce automated labeling~\citep{liu2025mmc, guan2025rstarmath, lightman2023lets}, most PRMs still rely on scalar rewards and struggle to generalize across tasks or training stages. We propose EduPRM, a process-aware reward model that critiques reasoning steps with scores, tags, and justifications. It integrates three complementary data strategies to capture multi-grained educational reasoning patterns, enabling scalable and interpretable reward modeling. EduPRM further serves as a unified backbone across data selection, trajectory search, and output reranking.


\subsection{MCTS-base LongCoT}
 LongCoT reasoning is essential for complex scientific tasks, yet methods like LLaVA-CoT often rely solely on prompting and struggle to produce coherent~\citep{xu2025llavacotletvisionlanguage}, diverse trajectories. Monte Carlo Tree Search (MCTS) provides a more structured alternative via iterative reasoning path expansion, but its direct application to MLLMs faces two challenges: ineffective search due to limited step-level supervision~\citep{ren2025deepseekproverv2, zhang2024restmcts,gao2024interpretable,zhang2024accessing}, and high computational cost in multi-agent or tool-augmented settings~\citep{br2024vermcts, yao2024mulberry}. We propose EduMCTS, a domain-adapted framework that introduces bootstrapping actions tailored to educational reasoning—such as a Self-Reflection mechanism for reflective error correction—and leverages fine-grained feedback from EduPRM to guide the search toward more accurate and pedagogically meaningful reasoning trajectories.

\section{EduFlow}

\subsection{EduPRM}

\subsubsection{EduPRM-420K}


We curate a corpus of 160K high-quality multimodal science problems from scanned exams, online platforms, and structured question banks. Images are parsed using Mathpix and LayoutLMv3, equations are LaTeX-encoded, and layouts are structured into JSON. After applying difficulty and text-solvability filtering, we extract 17K curriculum concepts via clustering and organize them into 54 competency groups to guide data selection and ensure comprehensive knowledge coverage.

To capture diverse reasoning patterns, we construct EduPRM-420K using three complementary strategies with task-specific supervision: (1) \textbf{MCTS-guided trajectories} (150K), where selected problems are processed by EduMCTS to generate verified, multi-step solutions through structured search; (2) \textbf{error-injected critiques} (150K), where GPT-4o-0513 modifies reference answers by inserting one of nine predefined error types into specific steps to simulate common misconceptions; and (3) \textbf{teacher–student dialogues} (120K), where lightweight models (e.g., Qwen2.5-VL-7B) produce initial responses, which are then reviewed and annotated by a stronger teacher model (Qwen2.5-VL-72B) with error types, explanations, and scores. These strategies jointly provide diverse supervision signals—from correct trajectories to realistic and noisy reasoning paths—forming a robust foundation for process-level reward modeling.

We define nine pedagogically motivated step-level labels to characterize student reasoning quality: Correct Step, Visual Misunderstanding, Problem Misunderstanding, Lack of Domain Knowledge, Misapplication of Knowledge, Logical Reasoning Error, Hallucination, Computational Error, and Off-topic or Incongruent. These categories cover a broad range of typical errors observed in educational scenarios.

Each instance in EduPRM-420K includes dual-level annotations: a sequence of structured reasoning steps, where each step is represented as a triple \textbf{\{content, label, explanation\}}. This enables supervision at both fine-grained and holistic levels.

To ensure annotation quality, we apply a rigorous filtering pipeline, see appendix.

Details in Figure~\ref{fig:prmpipeline}. After filtering, we retain 420K high-quality annotated examples, constituting the \textbf{EduPRM-420K} dataset for training.

\subsubsection{Training the PRM Model}

Our training data adopts two complementary supervision formats. The \textbf{Stepwise Format} provides a predefined list of reasoning steps, each annotated as a quadruple \{content, label, explanation, score\}. The \textbf{Critique Format} starts from a full student answer, which is first decomposed into steps and then annotated in the same quadruple form. This dual setup enables EduPRM to handle both structured step-level supervision and open-ended solution critique.

To support this, we adopt a \textbf{two-stage curriculum learning strategy}. The model is first trained on the Stepwise Format for localized reward prediction and error classification, then fine-tuned on the Critique Format for full-path assessment and explanation generation.

\begin{figure*}[t]
\centering
\includegraphics[width=1.0\textwidth]{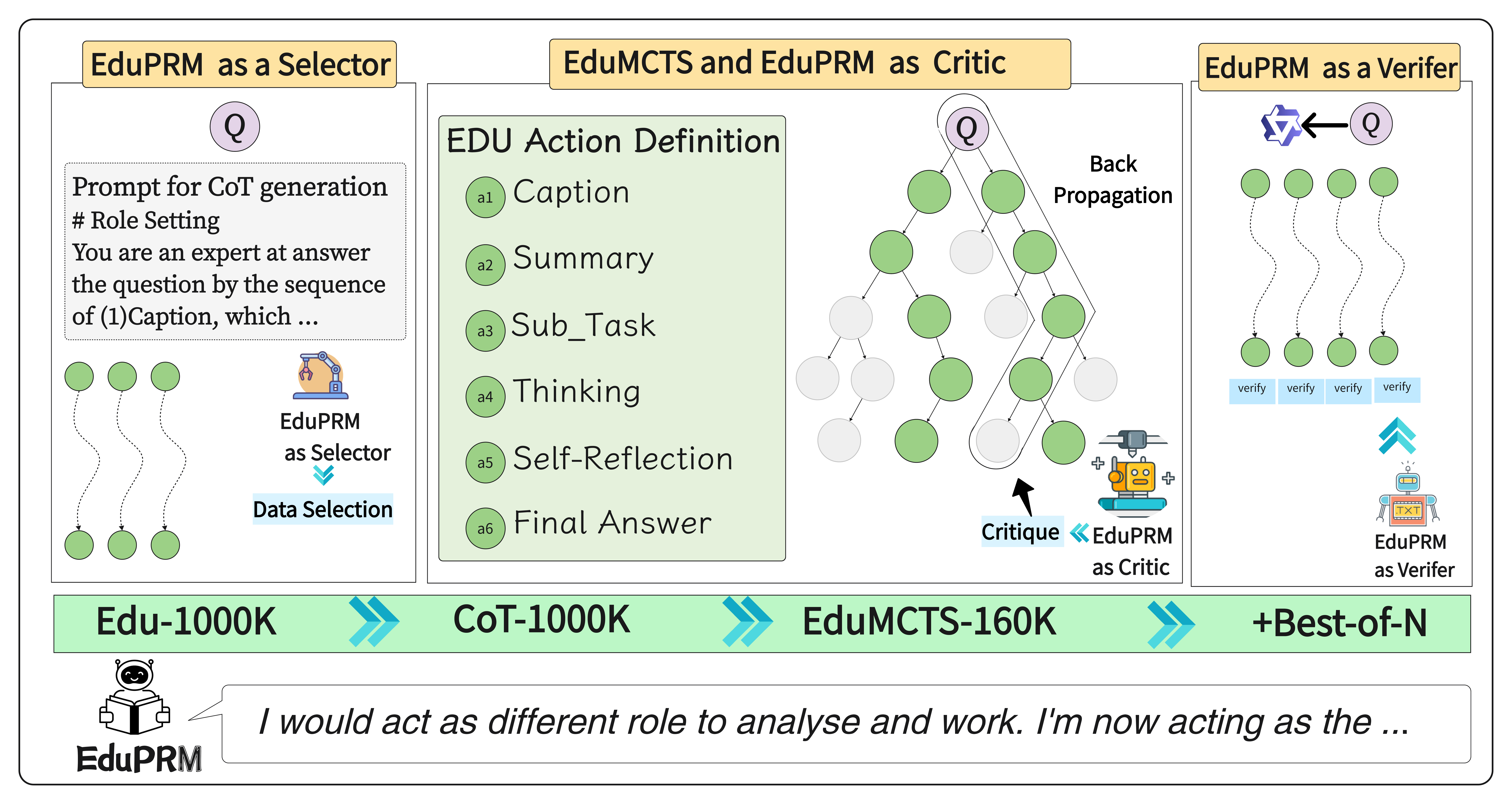} 
\caption{Overview of the multi-stage usage of \textbf{EduPRM} in the \texttt{EduFlow}. 
EduPRM plays distinct roles at each stage: 
(1) as a \textbf{Selector} to filter CoT outputs during \textit{Raw CoT Generation}, enabling data selection; 
(2) as a \textbf{critic} to evaluate and provide stepwise feedback on rollouts during \textit{EduMCTS Exploration}, guiding reward propagation and effectively search;
(3) as a \textbf{verifier} to select the best answer among sampled responses during \textit{BoN Inference}. 
}
  \label{fig:eduflow}
\end{figure*}

\subsection{Integrating EduPRM into EduFlow}

EduPRM functions as a unified process-level controller within EduFlow, enhancing both training and inference through fine-grained stepwise supervision. It is integrated at three key stages: (1) sample selection, (2) trajectory construction, and (3) inference-time reranking. Figure~\ref{fig:eduflow} illustrates the complete EduFlow loop, where EduPRM acts as a general-purpose evaluator across stages—from sample filtering and search guidance to output verification.

\subsubsection{PRM-Guided Sample Selection}

To improve training data quality, we apply EduPRM to identify ambiguous or under-explored samples from a candidate solution pool. Specifically, we evaluate each step in preliminary rollouts and compute reward variance across sampled solutions. Samples with high stepwise disagreement—often indicating ill-formed logic or under-learned behaviors—are prioritized for further exploration via MCTS. This allows EduFlow to selectively augment the most pedagogically valuable data.

\subsubsection{PRM-Guided MCTS Rollout}

During trajectory construction with MCTS, EduPRM is used as the reward function to evaluate intermediate states. At each rollout step, it scores partial trajectories based on stepwise coherence and correctness, which are then backpropagated through the search tree. This guides the search toward higher-quality reasoning paths and enables more efficient pruning of unpromising branches.

\subsubsection{PRM-Guided Inference and Reranking}

At inference time, EduPRM supports a lightweight test-time scaling (TTS) strategy. Multiple candidate outputs are sampled, and EduPRM scores their intermediate steps. The output with the highest accumulated stepwise reward is selected as the final answer. This approach improves output reliability while avoiding costly invocations of external models such as GPT-4V.





\subsection{EduMCTS}

EduMCTS constructs structured reasoning trajectories by combining action-specific search with process-level reward modeling. The core of EduMCTS is a decomposition of reasoning into six functionally distinct node types: \textbf{caption}, \textbf{summary}, \textbf{sub\_task}, \textbf{thinking}, \textbf{self-reflection}, and \textbf{answer}. These nodes reflect key components of problem-solving workflows, such as extracting visual information, abstracting task goals, decomposing subtasks, executing inference, verifying logic, and reporting conclusions, which are often critical in scientific reasoning.

Candidate reasoning paths are expanded from multiple MLLMs by sequentially sampling actions corresponding to the above node types. After each rollout, \textbf{EduPRM} evaluates the entire trajectory and assigns a reward based on step-level correctness, coherence, and the presence of identifiable errors. This score is then backpropagated to guide node selection in subsequent rollouts. Unlike conventional reward models that rely on scalar labels or outcome-only signals, EduPRM integrates error types, justifications, and reward intensity, allowing search to focus on logically sound and diagnostically informative trajectories.

To ensure output quality, we apply a two-stage filtering process. First, only trajectories in which all steps are labeled as correct are retained. Second, paths with low overall confidence or failed verification—especially during \textbf{Self-Reflection}—are discarded. This removes unstable or unsupported reasoning while preserving paths that exhibit internal consistency and meaningful error-checking behavior.

Applying EduMCTS to our raw dataset of multimodal science questions(see Appendix), we generate multiple rollouts per problem. After filtering, we obtain \textbf{EduMCTS-160K}, each sample paired with a multimodal input and a verified action sequence. This dataset supports models in learning not just accurate predictions, but reasoning strategies that are structured, verifiable, and aligned with scientific task requirements.

We fine-tune our MLLM on EduMCTS-160K. The model learns to generate step sequences that reflect consistent logic and structured problem-solving, informed by the search and filtering process guided by EduPRM.

\begin{table*}[ht]
\centering
\footnotesize
\begin{tabular}{lcccc|cccc|c}
\toprule
\multirow{2}{*}{\textbf{Method}} 
& \multicolumn{4}{c}{\textbf{Direct Inference Score}} 
& \multicolumn{4}{c}{\textbf{Step-by-Step Score}} 
& \multirow{2}{*}{\textbf{Avg.}} \\
\cmidrule(lr){2-5} \cmidrule(lr){6-9}
& Primary & Middle & High & Overall & Primary & Middle & High & Overall & \\ 
\midrule

\multicolumn{10}{c}{\textbf{\textit{Closed-Source Models}}} \\
GPT-4o & 45.56 & 37.42 & 30.39 & 35.02 & 48.28 & 37.44 & 29.80 & 35.00 & 35.01 \\
Gemini2-Flash & 56.70 & 55.17 & 45.69 & 51.08 & 54.63 & 51.12 & 41.95 & 47.34 & 49.21 \\
Gemini2-Thinking & 60.79 & 57.75 & 52.02 & 55.47 & 62.06 & 59.52 & 54.18 & 57.36 & 56.42 \\
\midrule

\multicolumn{10}{c}{\textbf{\textit{Open-Source Models}}} \\
Qwen2.5-VL-72B & 54.08 & 54.72 & 47.39 & 51.39 & 55.27 & 53.83 & 44.79 & 49.93 & 50.66 \\
Qwen2-VL-72B & 40.17 & 37.92 & 29.74 & 34.48 & 32.88 & 29.32 & 20.48 & 25.71 & 30.10 \\
InternVL2.5-MPO-78B & 50.32 & 48.22 & 41.55 & 45.43 & 50.23 & 46.15 & 37.86 & 42.82 & 44.13 \\
InternVL2.5-78B & 45.49 & 43.62 & 36.04 & 40.41 & 47.29 & 42.10 & 33.05 & 38.53 & 39.47 \\
InternVL2.5-MPO-8B & 33.12 & 33.41 & 25.68 & 29.94 & 35.37 & 30.46 & 21.53 & 26.93 & 28.44 \\
InternVL2.5-8B & 28.90 & 30.53 & 23.63 & 27.31 & 29.89 & 25.88 & 17.93 & 22.69 & 25.00 \\
LLaVA-OneVision-72B & 33.68 & 34.59 & 28.01 & 31.57 & 32.70 & 30.10 & 22.88 & 27.11 & 29.34 \\
\midrule

\multicolumn{10}{c}{\textbf{\textit{Baseline Variants}}} \\
Qwen2-VL-2B-Instruct & 19.13 & 20.15 & 15.98 & 18.20 & 8.25 & 7.57 & 4.74 & 6.38 & 12.29 \\
~~+ EduMCTS16W & 23.92 & 23.71 & 18.88 & 21.57 & 12.44 & 14.46 & 11.36 & 12.90 & 17.23 \\
~~+ EduMCTS16W + BoN & 24.84 & 24.32 & 20.65 & \textbf{23.00} & 14.39 & 16.83 & 14.70 & \textbf{15.84} & \textbf{19.42} \\
\midrule
Qwen2VL-7B-Instruct & 30.33 & 28.51 & 21.24 & 25.43 & 18.74 & 14.63 & 9.22 & 12.58 & 19.01 \\
~~+ EduMCTS16W & 38.21 & 36.49 & 29.42 & 33.04 & 34.68 & 32.24 & 25.45 & 29.42 & 31.23 \\
~~+ EduMCTS16W + BoN & 40.89 & 38.92 & 33.84 & \textbf{36.14} & 37.20 & 36.91 & 31.01 & \textbf{34.39} & \textbf{35.27} \\
\midrule
Qwen2.5VL-3B-Instruct & 34.39 & 36.46 & 26.88 & 31.99 & 25.54 & 18.98 & 11.98 & 16.45 & 24.22 \\
~~+ EduMCTS16W & 40.86 & 39.53 & 30.92 & 35.81 & 34.32 & 32.01 & 24.21 & 29.20 & 32.50 \\
~~+ EduMCTS16W + BoN & 41.88 & 41.67 & 33.01 & \textbf{37.85} & 37.22 & 35.78 & 28.86 & \textbf{33.42} & \textbf{35.64} \\
\midrule
Qwen2.5VL-7B-Instruct & 40.40 & 44.97 & 34.32 & 39.82 & 38.67 & 30.90 & 20.89 & 27.17 & 33.49 \\
~~+ EduMCTS16W & 49.66 & 48.17 & 38.59 & 44.03 & 44.88 & 43.03 & 34.78 & 39.50 & 41.77 \\
~~+ EduMCTS16W + BoN & 51.94 & 49.08 & 41.23 & \textbf{45.94} & 48.19 & 46.69 & 38.85 & \textbf{43.28} & \textbf{44.61} \\
\bottomrule
\end{tabular}
\caption{Comparison of model performance on the \textbf{K12Vista} benchmark under both Direct Inference and Step-by-Step evaluation settings. BoN, N=8.}
\label{tab:mainresultsavg}
\end{table*}

\section{Experiments}

\subsection{Settings}
To demonstrate the effectiveness and generalizability of our MCTS-based pipeline, we apply it to a family of Qwen models with varying sizes, including Qwen2.5-VL-7B, Qwen2-VL-7B, Qwen2-VL-2B, and Qwen2.5-VL-3B. We conduct comparisons against three categories of strong baselines across multiple multimodal reasoning benchmarks. Implementation details are provided in Appendix~\ref{sec:appendixExperimentsDetails}.

\subsection{Main Results}
We conduct comparative experiments on four Qwen-based models of different scales to evaluate the effectiveness of our proposed EduFlow pipeline. 
As shown in Table~\ref{tab:mainresultsavg}, across all model sizes, incorporating EduMCTS16W leads to substantial performance gains, and further applying BoN (N=8) inference consistently brings additional improvement. For example, Qwen2.5VL-7B-Instruct improves from a baseline score of 33.49 to 41.77 after EduMCTS16W fine-tuning, and further to 44.61 with BoN (N=8). 
Similar trends are observed for smaller models (e.g.Qwen2.5VL-3B, Qwen2VL-2B), demonstrating the scalability and generality of our method. Notably, our best-performing variant surpasses most state-of-the-art open-source and even closed-source models in both direct and step-by-step reasoning, highlighting the effectiveness of our methods.

In addition to K12Vista, we evaluate EduFlow on three challenging multimodal benchmarks—MDK12, MathVision, and MMMU-Pro-V-COT—as shown in Table~\ref{tab:leadboard}. 
Across all three datasets, EduFlow consistently improves accuracy over instruction-only baselines. 
For instance, Qwen2.5VL-7B improves from 26.87 to 30.46 average score, with gains observed on all three benchmarks (+3.62 on MDK12, +2.80 on MathVision, +4.35 on MMMU\_Pro\_V\_COT). 
These results demonstrate the cross-task generalization inprovements.

\begin{table}[t]
\centering
\footnotesize
\setlength{\tabcolsep}{3pt}
\definecolor{genRM}{RGB}{230,245,255}  
\definecolor{discRM}{RGB}{255,235,210} 
\definecolor{top1}{gray}{0.3}
\definecolor{top2}{gray}{0.6}
\definecolor{top3}{gray}{0.85}

\begin{tabular}{lccc|c}
\toprule
&MDK12 &MV &MMMU. &Avg. \\
\midrule
Qwen2-VL-2B-Inst &22.97&12.4&11.04	&15.47\\
~~~~+ EduMCTS16W &24.17&15.13&18.94	&19.41 \\
\midrule
Qwen2VL-7B-Inst &29.07&16.3&14.39	&19.92 \\
~~~~+ EduMCTS16W &31.77&20.21&17.69	&23.22 \\
\midrule
Qwen2.5VL-3B-Inst &30.13&20.06&13.87	&21.35 \\
~~~~+ EduMCTS16W &34.06&22.15&16.61	&24.27 \\
\midrule
Qwen2.5VL-7B-Inst &38.24&25.49&16.88	&26.87 \\
~~~~+ EduMCTS16W &41.86&28.29&21.23	&30.46 \\
\bottomrule
\end{tabular}
\caption{Performance of our EduFlow on Qwen models on three multimodal benchmarks: \textbf{MDK12}, \textbf{MathVison}, and \textbf{MMMU_Pro_V_COT}. }
\label{tab:leadboard}
\end{table}

\subsection{Contributions of Each Stage}
Figure~\ref{fig:ablationstages} shows the contribution of training and inference stage in the EduFlow pipeline. 
We report step-by-step reasoning scores for four Qwen variants across five progressive configurations: Baseline, Edu-1000K, CoT-1000K, EduMCTS-160K, and BoN (N=8).
Compared to supervised fine-tuning on Edu-1000K or CoT-1000K, which results in limited gains or even degradation in some cases, the introduction of EduMCTS-160K consistently leads to large improvements across all model sizes. 
For example, Qwen25VL-7B improves from 22.23 (after CoT-1000K) to 39.50 with EduMCTS training, and further increases to 43.28 with BoN-based inference. 
Smaller models such as Qwen25VL-3B also benefit significantly, with step-by-step scores rising by 21.81 points over their respective baselines.
These results confirm that each component of the EduFlow pipeline contributes cumulatively to model performance, and that the improvements are robust across different model size.

\section{Discussion and Analysis}

We perform comprehensive ablation studies to understand the contributions of our key components, including PRM, BoN inference, MCTS pipeline configuration, and subject-wise performance.

\subsection{EduPRM Validation}
Table~\ref{tab:ablationqwen25} presents the evaluation of various verifier models on K12-PEBench and their effectiveness in guiding inference under the K12Vista benchmark with BoN (N=8). We compare six verifier candidates, including general-purpose vision-language models (e.g., Qwen2-VL-7B), and our EduPRM (7B). Among models of similar scale, EduPRM achieves the highest accuracy on K12-PEBench (54.69), indicating its stronger step-level verification capability. Compared to Qwen2-VL-72b, which attains a slightly higher verification score (54.89), EduPRM delivers better downstream selection accuracy on K12Vista (42.23 vs. 40.13), suggesting that EduPRM is better aligned with educational reasoning quality beyond general token-level preferences.

Furthermore, as shown in Table~\ref{tab:bon}, EduPRM demonstrates superior scalability in Best-of-N inference. While other strategies exhibit saturation or even performance degradation when increasing $N$ from 4 to 8, EduPRM continues to improve consistently—from 41.21 (N=2) to 43.28 (N=8) and 43.45 (N=12)—confirming its robustness in selecting high-quality completions from larger candidate pools.

\begin{table}[ht]
\centering
\footnotesize
\setlength{\tabcolsep}{4pt}
\begin{tabularx}{\linewidth}{Xcc}
\toprule
\multirow{2}{*}{\textbf{Verify Models}} 
& \textbf{K12-PEBench} & \textbf{K12Vista} \\
& \textbf{ACC}         & \textbf{(BoN, N=4)} \\
\midrule
Qwen2-VL-7B                      & 42.76 & 33.50 \\
InternVL25-8B                   & 46.96 & 38.09 \\
Qwen25-VL-7B-Inst               & 48.74 & 37.28 \\
InternVL25-MPO-8B              & 53.19 & 38.61 \\
Qwen2-VL-72b                   & 54.89 & 40.13 \\
EduPRM(7B) (\textbf{Ours})      & \textbf{54.69} & \textbf{42.23} \\
\bottomrule
\end{tabularx}
\caption{Evaluation of various verifier models}
\label{tab:ablationqwen25}
\end{table}

\begin{table}[ht]
\centering
\footnotesize
\setlength{\tabcolsep}{5pt}
\begin{tabularx}{\linewidth}{Xcccc}
\toprule
\textbf{Method} & \textbf{N=2} & \textbf{N=4} & \textbf{N=8} & \textbf{N=12} \\
\midrule
Random Sample      & 38.87 & 39.93 & 39.61 & 39.12 \\
Self Consistency   & 39.50 & 40.02 & 40.62 & 41.30 \\
Qwen2.5vl-7B-ORM   & 38.89 & 40.01 & 36.24 & 39.22 \\
Qwen2.5vl-7B-PRM   & 39.06 & 38.94 & 38.00 & 39.07 \\
EduPRM             & 41.21 & 42.23 & \textbf{43.28} & 43.45 \\
\bottomrule
\end{tabularx}
\caption{Accuracy on the \protect\textbf{K12Vista} benchmark under different sampling sizes \(N \in \{2, 4, 8, 12\}\), with baseline \(N=1\) performance 39.50\%.}
\label{tab:bon}
\end{table}

\subsection{Impact of MCTS Pipeline Design}

Tables~\ref{tab:ablationmcts1} and~\ref{tab:successraterefined} evaluate the impact of core components in our EduMCTS, including data selection, structured action control, reward modeling, and rollout strategy. Starting from the baseline Qwen25-vl-7b, directly applying MCTS with a weak verifier results in degraded performance (36.13 direct, 25.64 step-by-step), primarily due to unfiltered search expansions. Incorporating data selection (+2.82 step) helps reduce low-quality branches, while stepwise action modeling further improves structure-aware rollouts (+3.81 step), leading to more coherent paths.
Introducing EduPRM as a value critic adds significant gains (37.69 step), by providing learning-aligned, process-level supervision to guide backpropagation. When paired with rollout-based voting (N=4), performance peaks at 39.50 step, and further improves with N=8.
Table~\ref{tab:successraterefined} shows that success rate improves from 67.8\% (vanilla MCTS) to 73.3\% (+actions) and 87.1\% (+EduPRM), indicating better convergence with fewer invalid or redundant paths.

\begin{table}[ht]
\centering
\footnotesize
\setlength{\tabcolsep}{4pt}
\begin{tabularx}{\linewidth}{lcc}
\toprule
\textbf{Method} & \textbf{Direct } & \textbf{Step} \\
\midrule
Baseline (Qwen25-vl-7b) & 39.82 & 27.17 \\
\hspace{0.5em}+MCTS (Judge:Qwen25-vl-7b,Ro:1) & 36.13 & 25.64 \\
\hspace{1em}+Data Selection & 38.95 & 28.43 \\
\hspace{1.5em}+Stepwise Action Nodes & 40.10 & 32.24 \\
\hspace{2em}+EduPRM Judge & 42.01 & 37.69 \\
\hspace{3em}+Rollouts (2) & 42.81 & 38.14 \\
\hspace{3em}+Rollouts (4, \textbf{Ours}) & \textbf{44.03} & \textbf{39.50} \\
\hspace{3em}+Rollouts (8) & 44.08 & 39.71 \\
\bottomrule
\end{tabularx}
\caption{Ablation results under Direct and Step-by-Step settings for Qwen25-vl-7b. Each row adds a component or strategy incrementally.}
\label{tab:ablationmcts1}
\end{table}

\begin{table}[ht]
\centering
\footnotesize
\renewcommand{\arraystretch}{1.1}
\begin{tabularx}{\linewidth}{>{\raggedright\arraybackslash}p{5.2cm} c}
\toprule
\textbf{Method} & \textbf{Success Rate} \\
\midrule
MCTS (Judge: Qwen25-vl-7b) & 67.8 \\
\addlinespace[0.3ex]
\hspace{1em}+ Stepwise Action Nodes & 73.3 \\
\hspace{2em}+ EduPRM Judge (\textbf{Ours}) & \bfseries 87.1 \\
\bottomrule
\end{tabularx}
\caption{Progressive improvement in \textbf{success rate} through stepwise and reward-based supervision.}
\label{tab:successraterefined}
\end{table}

\subsection{Effect of N in Best-of-N Inference}
Table~\ref{tab:bon} presents the effect of increasing the number of sampled reasoning paths ($N \in \{2, 4, 8, 12\}$). Notably, we observe a strong upward trend from baseline to $N=8$, where accuracy peaks across most categories. However, the improvements between $N=8$ and $N=12$ become marginal.

\subsection{Performance Across Subjects}
\begin{figure}[ht]
  \centering
    \includegraphics[width=0.9\linewidth]{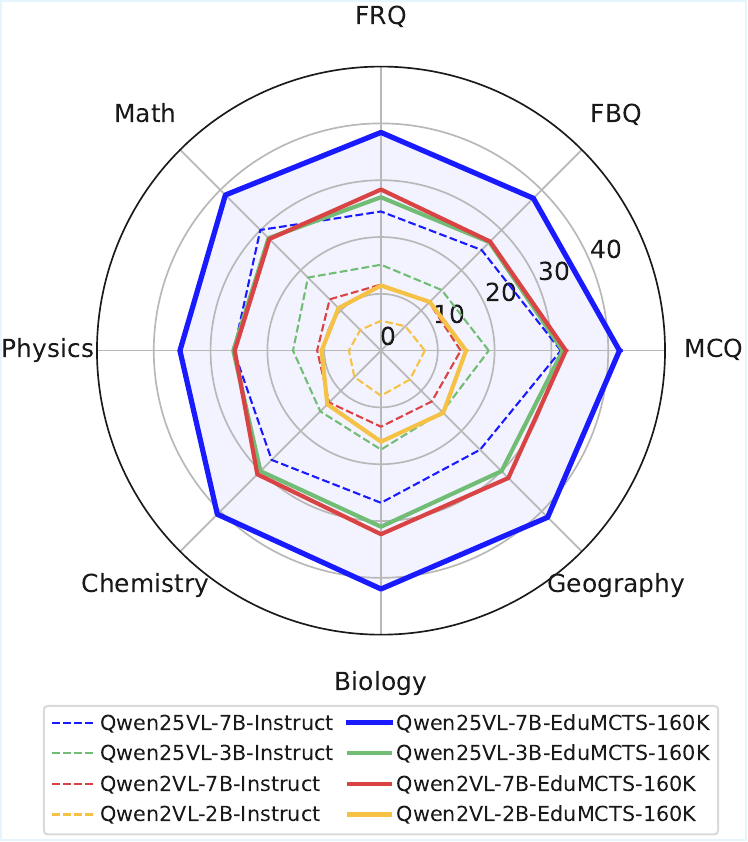}
  \caption{\textbf{Subject and question type specific 
}accuracy. STEM domains see marked improvement under our method.}
  \label{fig:graderadar}
\end{figure}

Figure~\ref{fig:graderadar} presents subject-wise accuracy comparisons across core K-12 domains. Our method consistently improves performance across all subjects, with particularly significant gains in STEM fields such as Math, Physics, and Chemistry.

\section{Conclusion}
EduFlow represents a significant advancement in optimizing scientific reasoning. By integrating MCTS-guided data construction, supervised fine-tuning, and best-of-N selection, EduFlow offers a comprehensive pipeline that enhances educational reasoning processes. Central to this framework is EduPRM, the pioneering process reward model tailored for educational reasoning across diverse scenarios and stages. The development of EduMCTS, an innovative Monte Carlo Tree Search algorithm, further enriches this pipeline by incorporating structured educational actions and self-reflection mechanisms. This has facilitated the creation of EduMCTS-160K, a substantial dataset comprising 160K multimodal science problems with intricate reasoning trajectories. Our extensive experiments across K–12 benchmarks, supported by thorough ablation studies, demonstrate that EduFlow achieves excellent results, particularly excelling in process-oriented evaluation metrics. These findings underscore the potential of EduFlow to transform educational reasoning.

\newpage
\section{Limitations}
\subsection{Experimental Setup}
The base model was tested only on Qwen2VL and Qwen25VL, without testing on additional base models.
The base model was tested only on Qwen2VL and Qwen25VL, without testing on additional base models. Furthermore, the experimental domain was limited to the Edu field, without expansion to other domains; the main experimental results focused on the K12 education sector, specifically K12Vista and MDK12, without broader attention to other domain benchmarks. Additionally, EduMCTS lacks comparison with more MCTS methods, including additional versions of the Value model and the Policy model. Finally, the BoN experiments did not further test results under settings such as N=128, N=256, etc.

\subsection{Limited Exploration of Reasoning Models}
At present, deep reasoning models such as R1 consistently deliver impressive outcomes. Nonetheless, there is insufficient comprehensive research into these models, especially regarding the elements that improve their ability to follow instructions.

\newpage
\newpage
\bibliography{custom}

\appendix

\section{Appendix}
\label{sec:appendix}

\subsection{Experiments Details}
\label{sec:appendixExperimentsDetails}

\textbf{Frontier LLMs:} GPT-4o, Claude (latest version), OpenAI o1-preview, and o1-mini. Benchmark scores for these models are taken from official technical reports [Team, 2024a].

\textbf{Open-sourced superior reasoning models:} DeepSeek-Coder-v2-Instruct, Mathstral [Team, 2024b], NuminaMath-72B [Jia Li and Polu, 2024a], LLaMA3.1 [Dubey et al., 2024].

\textbf{Instruct and RM-tuned variants:} Qwen2.5-Math-7B-Instruct, Qwen2.5-Math-72B-Instruct + Qwen2.5-Math-RM-72B. Notably, these models use a 72B reward model (ORM), significantly larger than our 7B PRM.~\citep{guo2025deepseek,wang2024enhancing,Qvq,Gpt-o3-mini,Gpt-o1-mini,gpt4v,gpt4,qwen25,deepseek-r1,li2024llava,Internal25vl,yang2024qwen2,Claude}

\paragraph{Benchmarks}To better explore the ability on science subjects, we mainly evaluate on subejects process supervision benchmarks: K12PeBench, designed for evaluating reasoning process supervision across disciplines, We report both direct final accuracy and step-by-step intermediate accuracy for each benchmark to comprehensively assess reasoning quality.

Besides above, we evaluate our approach on a variety of multimodal reasoning benchmarks, encompassing both general-purpose and science-specific datasets.

\paragraph{Implementation Details}
For the SFT stage, we use the AdamW optimizer with a learning rate of \( 2.0 \times 10^{-6} \) and a batch size of 4. During inference for all baselines, we set the decoding temperature to 0 or a near-zero value to ensure stable and deterministic outputs. For the Best-of-\(N\) decoding strategy, we sample 8 candidate responses with a temperature \( T \sim \mathrm{Uniform}(1.1, 1.3) \). Evaluation is conducted using \texttt{VLMEvalKit} and official benchmark codebases.

\subsection{MCTS}
\begin{itemize}
    \item \textbf{(a) Expansion:} 
    Given the current node \( s_t \), we use a pool of actor models \( \{ \pi_1, \ldots, \pi_K \} \) to generate diverse candidates:
    \[
        \mathcal{C}_t = \bigcup_{j=1}^{K} \pi_j\left( \cdot \mid s_t \right)
    \]
    where \( \mathcal{C}_t \) is the set of all candidate nodes, and \( \pi_j \) is the \( j \)-th actor model.

    \item \textbf{(b) Scoring and Filtering:} 
    Each candidate \( s' \in \mathcal{C}_t \) is scored by EduPRM:
    \[
        \hat{r}(s') = \frac{1}{K} \sum_{j=1}^K \mathrm{PRM}_\psi\left( \textbf{Yes} \mid s' \right)
    \]
    where \( \hat{r}(s') \) is the averaged stepwise reward, and \( \mathrm{PRM}_\psi \) denotes the process reward model. Low-scoring steps are discarded:
    \[
        \mathcal{C}_t^+ = \left\{ s' \in \mathcal{C}_t \mid \hat{r}(s') \geq \tau \right\}
    \]
    where \( \tau \) is a fixed pruning threshold.

    \item \textbf{(c) Backpropagation:}
    For each ancestor node \( s \), we update:
    \[
        V(s) \leftarrow \frac{N(s) \cdot V(s) + \sum \hat{r}(s')}{N(s) + |\mathcal{C}_t^+(s)|}
    \]
    \[
        N(s) \leftarrow N(s) + |\mathcal{C}_t^+(s)|
    \]
    where \( V(s) \) and \( N(s) \) denote the value and visit count of \( s \), and \( \mathcal{C}_t^+(s) \) denotes surviving children of \( s \).

    \item \textbf{(d) Selection:}
    The next node is selected via UCB:
    \[
        s_{t+1} = \arg\max_{s \in \mathcal{C}_t^+} \left[ V(s) + c \cdot 
        \sqrt{ \frac{\log N(\mathrm{parent}(s))}{1 + N(s)} } \right]
    \]
    where \( c \) is the exploration coefficient.
\end{itemize}

This four-step loop repeats until a complete reasoning trajectory is constructed or a maximum search depth is reached.

We extract 17K curriculum-aligned concepts using a BERT-based model and cluster them into 54 competency groups. Stratified sampling follows:
\[
w_{s,g} = \frac{N_{s,g}}{\sum_{s=1}^5 \sum_{g=7}^{12} N_{s,g}} \times T
\]
where \(T=160{,}000\), \(N_{s,g}\) is the number of concepts in subject \(s\), grade \(g\). To ensure challenge and visual grounding, we remove questions easily solved by LMMs (>70\% acc.) or solvable without images.

\subsection{PRM step-level labels}
We define nine pedagogically informed step-level labels that reflect common reasoning errors in educational scenarios, derived from real-world student behavior and aligned with instructional goals:

\begin{itemize}
    \item \textbf{Correct Step}: No identifiable error among the types below.

    \item \textbf{Visual Misunderstanding}: Errors in interpreting images, spatial relationships, or diagrams—frequent in visually grounded tasks where students misread axes, misjudge geometric configurations, or extract incorrect values.

    \item \textbf{Problem Misunderstanding}: Misinterpretation of the question’s intent, constraints, or key information—reflecting difficulties in reading comprehension and requirement identification.

    \item \textbf{Lack of Domain Knowledge}: Gaps in subject knowledge or failure to recall relevant facts—common among students unfamiliar with specific scientific or mathematical content.

    \item \textbf{Misapplication of Knowledge}: Incorrect use of known concepts, formulas, or procedures—highlighting partial understanding or mechanical misuse.

    \item \textbf{Logical Reasoning Error}: Broken or invalid reasoning chains—manifesting in unsupported conclusions, skipped inference steps, or misused premises.

    \item \textbf{Hallucination}: Fabricated or logically incoherent statements, often arising when students infer beyond evidence or insert unrelated content.

    \item \textbf{Computational Error}: Arithmetic or algebraic mistakes—among the most frequent error types in student math reasoning.

    \item \textbf{Off-topic or Incongruent}: Step content mismatched with its functional intent (e.g., no visual description in a caption step), indicating a misalignment between task structure and student response.
\end{itemize}

\subsection{PRMdata Filtering Strategy}
\textbf{Format Consistency:} We remove samples that deviate from the required format, where each reasoning step must be annotated as a quadruple and the number of annotated steps must match the step sequence length.

\textbf{Annotation Consistency:} We verify consistency between step-level annotations from multiple models. Samples with inconsistent labels across annotators are discarded.

\textbf{Error Coverage:} We promote diversity by filtering out samples with overly repetitive error patterns or insufficient label variety.

\subsection{PRM prompt}

\begin{figure*}[h]
\centering
\includegraphics[width=1.0\textwidth]{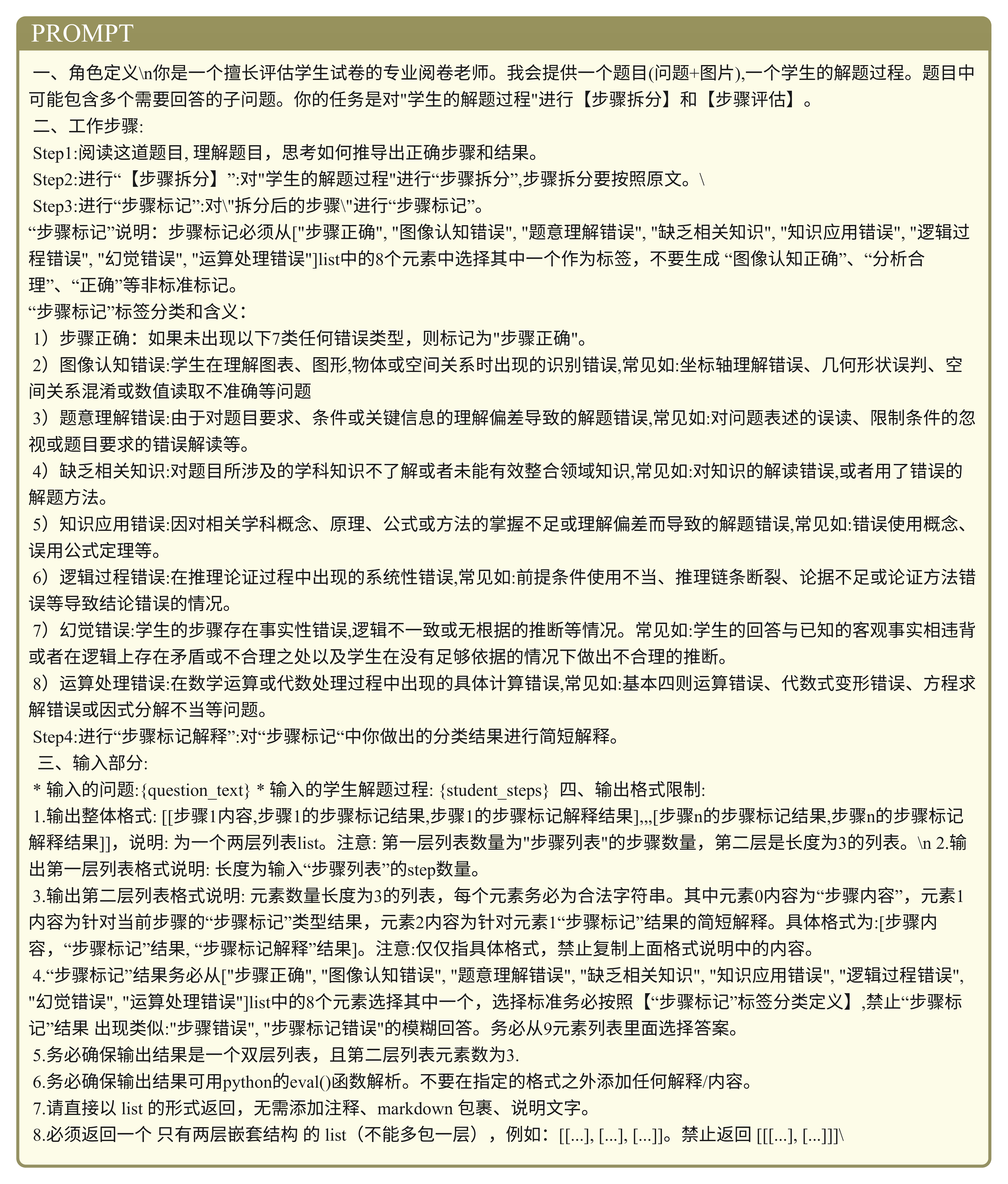} 
\caption{Prompt for step breakdown and annotation
}
\label{Factors}
\end{figure*}

\begin{figure*}[h]
\centering
\includegraphics[width=1.0\textwidth]{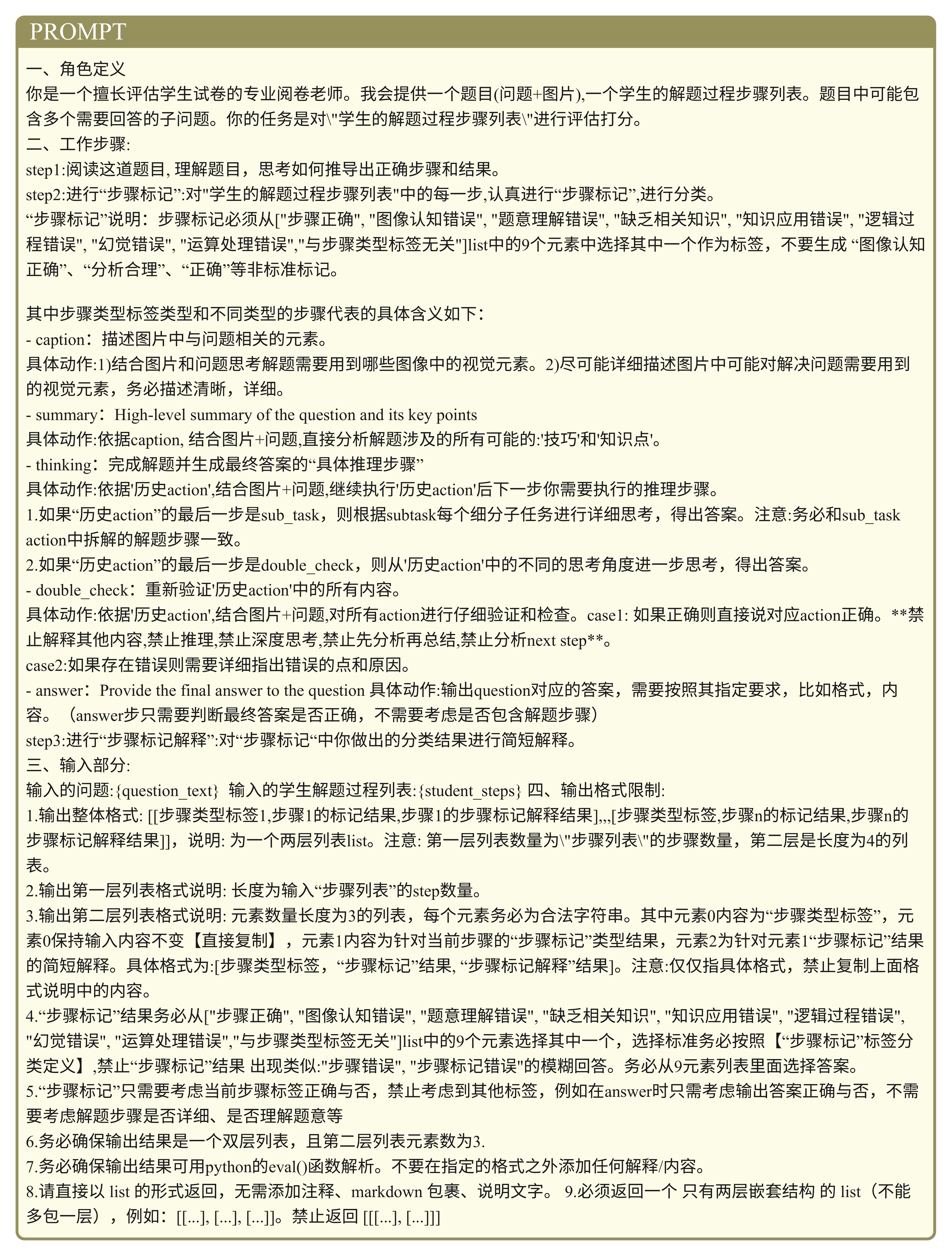} 
\caption{Prompt for step annotation
}
\label{Factors}
\end{figure*}

\end{document}